# Estimation and Deconvolution of Second Order Cyclo-Stationary Signals

Igor Makienko[a], Michael Grebshtein[a], Eli Gildish[a] *

[a]*RSL Electronics LTD, Migdal Ha'Emek, 23100, Israel*

**Abstract**

This method solves the dual problem of blind deconvolution and estimation of the time waveform of noisy second-order cyclo-stationary (CS2) signals that traverse a Transfer Function (TF) en route to a sensor. We have proven that the deconvolution filter exists and eliminates the TF effect from signals whose statistics vary over time. This method is blind, meaning it does not require prior knowledge about the signals or TF. Simulations demonstrate the algorithm's high precision across various signal types, TFs, and Signal-to-Noise Ratios (SNRs). In this study, the CS2 signals family is restricted to the product of a deterministic periodic function and white noise. Furthermore, this method has the potential to improve the training of Machine Learning models where the aggregation of signals from identical systems but with different TFs is required.

## 1. Introduction

Cyclo-stationary signals, which are the focus of this study, exhibit periodic variations in their statistical properties. Specifically, in wide-sense second-order cyclo-stationary (CS2) signals, the first two moments change periodically [1]. These signals are prevalent in numerous domains, including telecommunications, telemetry, radar, sonar, mechanics, radio astronomy, econometrics, and atmospheric science [2]. In the field of mechanics, the rotation of machinery is a significant source of such periodicity. Early signs of faults in gears, bearings, or components of internal combustion engines are represented by CS2 signals, typically detected using vibration, acoustic, or pressure sensors [3]. In telecommunications, telemetry, radar, and sonar, the periodicity in statistics stems from processes such as modulation, sampling, multiplexing, and coding. In radio astronomy, periodicity is observed due to phenomena like planetary revolution, rotation, and star pulsations. Econometrics encounters periodicity induced by seasonality, while atmospheric science studies periodic variations resulting from the Earth's rotation and revolution [4].

In the literature, two types of CS2 detectors are distinguished: one that identifies the presence of a CS2 signal amidst noise [5], and another that estimates CS2 signals, assuming prior knowledge about the cycle period or signal's sparsity [6]. However, in real-world situations, this information might not be available. Furthermore, noisy CS2 signals often traverse a Transfer Function (TF), such as a communication channel in telecommunications, the atmosphere in astronomy, the earth in geophysics, or the system structural response in mechanics. This TF distorts the original information by significantly altering the signal amplitude at various frequencies [7]. The impact of the TF becomes even more critical in Machine Learning (ML) models, where data from identical physical systems with differing TFs should be incorporated into the model training process [8]. For instance, eliminating TF and CS2 waveform estimation plays a crucial role in data preprocessing for ML training in mechanical health monitoring applications [9],[10]. This is particularly important when historical data from identical machines is used to predict the remaining useful life of a machine [11], provide ML-based machine health monitoring [12] or extract valuable information from complex vibration signals using ML methodologies [13],[14].

Hence, there is a need for estimators that can not only detect the presence of CS2 signals but also estimate their waveform after the deconvolution eliminating the TF effect. This would enable the use of ML models that aggregate data from different systems of the same type. This study augments the current body of research by concurrently addressing two issues: executing deconvolution of the CS2 signal which traverses the TF en route to the sensor and estimating its time waveform. The signals examined in this study are constrained by the multiplication of a deterministic periodic function of interest and white noise, which is a common occurrence in many practical applications.

The paper is structured as follows: Section 2 presents the related works in this field; Section 3 provides all the definitions and outlines the problem; Section 4 explains the method and provides a summary of the algorithm; Section 5 showcases evaluation results using simulations and discusses the selection of the algorithm's parameters; Section 6 concludes the paper and offers recommendations for future research.

## 2. Related works

Several methodologies have been formulated to estimate signals that are influenced by TF, with a subset of them concentrating on cyclostationary signals. The most frequently used technique is the minimum entropy deconvolution (MED), first proposed in [15]. This algorithm aims to reduce the signal entropy, which is closely related to the signal's impulsive behavior, leading to a reduction in entropy value. Given that kurtosis is employed to emphasize spikes, the deconvolution is structured by maximizing kurtosis as a criterion. However, MED-based techniques have their limitations as they assume the signal of interest to exhibit impulsive characteristics.

A new non-iterative algorithm employs a novel indicator known as D-Norm, which is an enhanced version of MED, referred to as MEDD [16]. This enhancement improves the performance of MED in signals with a low Signal-to-Noise Ratio (SNR). Although this method is more computationally efficient, it necessitated matrix inversion. The computational efficiency was further augmented in [17] by avoiding the need for matrix inversion. Another deconvolution method uses an objective function based on a skewness-kurtosis criterion as introduced in [18]. Additionally, the $l_0$-norm and its hyperbolic tangent transformation were employed to enhance the sensitivity of the MED-based algorithms to outliers in the signal [19]. Most methods based on MED eliminate the TF effect under the assumption that the filtered signal should exhibit impulsive behavior, which is apt for certain applications. However, this is not ideal in our scenario where the filtered signal is presumed to represent a periodically modulated white noise.

In certain scenarios, the computation of the deconvolution filter is predicated on the assumption of signal sparsity. A new sparsity metrics, coupled with the Rayleigh optimization, have been employed in the design of a new deconvolution filter, as elucidated in [20] and [21].

An alternative set of methods concentrates on the deconvolution filters, which emphasize the cyclostationarity measure of the envelope spectrum. The optimization process is usually facilitated by a gradient descent algorithm [22]. The maximum kurtosis criterion, as utilized in MED, is suboptimal for cyclostationarity signals. Contrarily, some methods prioritize the maximization of the cyclostationarity measure of the signal over kurtosis in the design of the deconvolution filter, leading to what is known as second-order cyclostationarity blind deconvolution (CYCBD) [23]. In this scenario, the optimization problem is tackled iteratively using the eigenvalue decomposition algorithm.

Numerical techniques of the deconvolution filter design have been published in the realm of machinery fault diagnostics and were recently consolidated in [24]. A subsequent work [7], expanded on this summary with a focus on the surveillance of mechanical faults in gears.

In conclusion, the majority of methods in this field concentrate on the design of the deconvolution filter, making assumptions about the specific properties of the target signal, such as the existence of periodic impulses or signal sparsity. This significantly restricts the applicability of these methods. Furthermore, the literature has largely overlooked the resolution of the dual problem of deconvolution and estimation of the cyclostationary second-order (CS2) signal waveform, especially when no information about the signals or TF properties is available.

Our research contributes to the existing body of knowledge by addressing the dual problem of deconvolution and CS2 waveform estimation in the absence of information about the signals or the system TF properties. It's important to note that the CS2 signals considered in this study are restricted to white noise modulated by any periodic deterministic signal, as outlined in the paper.

## 3. Problem Statement

### 3.1. Definition1

Consider $c(t)$, a real-valued continuous-time signal. It is deemed to be second-order cyclostationary in the wide sense (CS2) with a period T>0 if its mean and autocorrelation function are periodic functions of time with a period T for all $t$ and $\tau$:

$$R_{cc}(t,\tau) \equiv \mathbb{E}\{c(t+\tau)c(t)\} = R_{cc}(t+T,\tau) \tag{1}$$

$$\mathbb{E}\{c(t+T)\} = \mathbb{E}\{c(t)\},$$

where $\mathbb{E}\{*\}$ denotes an expectation operator and $\tau$ represents the time lag between distinct instances.

### 3.2. Definition2

In our study, we utilize a specific CS2 signal, which is the product of a non-negative periodic deterministic signal, denoted as $q(t)$, and a zero-mean white noise signal, represented as $r(t)$:

$$c(t) = q(t)r(t). \tag{2}$$

Its mean and auto-correlation are equal to:

$$\mathbb{E}\{c(t)\} = q(t)\underbrace{\mathbb{E}\{r(t)\}}_{\equiv 0}=0 \tag{3}$$

$$R_{cc}(t,\tau) = \mathbb{E}\{c(t)c(t-\tau)\} = \begin{cases} q(t)^2\sigma_r^2, & \tau = 0 \\ 0, & \tau \neq 0 \end{cases} \tag{4}$$

where $\sigma_r^2$ is the variance of $r(t)$. For all values of $\tau$, except when $\tau = 0$ the function is zero because the carrier of $c(t)$ is white noise. When $\tau = 0$ the function equals the signal's instantaneous power, which varies periodically over time, given that $q(t)$ is, by definition, a periodic function. Consequently, the process $c(t)$ has a zero mean and is uncorrelated.

As we proceed with our study, we will consider $c(t)$ to be a CS2 signal. Our objective is to estimate the deterministic function $q(t)$.

### 3.3. Problem Statement

Let's define a real measurement signal, denoted as $x(t)$, as a superposition of two sources. These sources pass through a linear, invertible TF before reaching the sensor. The measured values are a linear combination of n past values of these sources, each multiplied by the corresponding coefficients of the TF. The sources include a CS2 signal, $c(t)$, as defined in 3.2, and a zero-mean white noise, $w(t)$:

$$x(t) = \sum_{i=1}^{n} h_i y(t-i) \tag{5}$$

$$y(t) = c(t) + w(t),$$

where:
$n$ is the length of the TF,
$c(t) = q(t)r(t)$ is the CS2 signal whose properties are defined in 3.2.
$w(t)$ is a white measurement noise not correlated with $c(t)$.

The objective of this study is to estimate the envelope, denoted as $q(t)$, using the measurement signal $x(t)$. We operate under the assumption that the TF is invertible, meaning its inverse exists. Consequently, we address the dual problem of performing blind deconvolution to eliminate the effect of the TF, and estimating the envelope of the CS2 signal.

## 4. Methodology

*4.1. Deconvolution Filter Estimation*

The signal $x(t)$ incorporates the CS2 component $c(t)$. As a result, the elements of the $n \times n$ correlation matrix, $\boldsymbol{R}_{xx}(t,\tau)$, exhibit time-variant characteristics:

$$\boldsymbol{R}_{xx}(t,\tau) = \begin{bmatrix} \sigma_x^2(t) & \cdots & r_x(t, 1-n) \\ \vdots & \ddots & \vdots \\ r_x(t, n-1) & \cdots & \sigma_x^2(t) \end{bmatrix}, \tag{6}$$

where $r_x(t,\tau) = \mathbb{E}\{x(t)x(t-\tau)\}$ and $\sigma_x^2(t)$ is the time-varying variance of $x(t)$. Due to the TF effect $\boldsymbol{R}_{xx}(t,\tau)$ is not diagonal and the off-diagonal elements are time-dependent.

Let's demonstrate that the time dependence of any element in $\boldsymbol{R}_{xx}(t,\tau)$ is determined solely by the variance of the CS2 component, denoted as $\sigma_c^2(t)$:

$$\begin{aligned} r_x(t,\tau) &= \mathbb{E}\{x(t)x(t-\tau)\} \\ &= \mathbb{E}\left\{\sum_{i=1}^{n} h_i\big(c(t-i) + w(t-i)\big) \sum_{j=1}^{n} h_j\big(c(t-j-\tau) + w(t-j-\tau)\big)\right\} \\ &= \sum_{i=1}^{n}\sum_{j=1}^{n} h_i h_j [\mathbb{E}\{c(t-i)c(t-j-\tau)\} + \mathbb{E}\{w(t-i)w(t-j-\tau)\}] = \\ &= (\sigma_c^2(t) + \sigma_w^2) \sum_{i-j=\tau} h_i h_j \equiv (\sigma_c^2(t) + \sigma_w^2)\Phi(\tau), \end{aligned} \tag{7}$$

where function $\Phi(\tau)$ depends on $\tau$ only and is defined as follows:

$$\Phi(\tau) \equiv \sum_{i-j=\tau} h_i h_j, \quad \forall i, j \in [1, n]. \tag{8}$$

Thus, in our scenario $\boldsymbol{R}_{xx}(t,\tau)$ can be reformulated as a product of the time-dependent expression $\sigma_c^2(t) + \sigma_w^2$ and a matrix $\boldsymbol{\Phi}(\tau)$. The elements of this matrix, denoted as $\Phi(\tau)$, represent multiplications between different TF coefficients and depend solely on $\tau$. This implies that $\boldsymbol{R}_{xx}(t,\tau)$ is periodic in time, with the same period $T$ as $\sigma_c^2(t)$:

I.Makienko, M. Grebshtein and E. Gildish

$$\boldsymbol{R}_{xx}(t,\tau) = (\sigma_c^2(t) + \sigma_w^2) \begin{bmatrix} \Phi(0) & \cdots & \Phi(1-n) \\ \vdots & \ddots & \vdots \\ \Phi(n-1) & \cdots & \Phi(0) \end{bmatrix} \equiv (\sigma_c^2(t) + \sigma_w^2)\boldsymbol{\Phi}(\tau) \quad (9)$$

The deconvolution filter should remain constant to correspond to the inverse of the constant TF. Let's prove this proposition. The function $\sigma_c^2(t)$ can be expressed as the sum of its constant average value $\bar{\sigma}_c^2$ and a zero-mean time-varying component $\tilde{\sigma}_c^2(t)$. We aim to demonstrate that the deconvolution filter, estimated using $\bar{\sigma}_c^2$ alone, should be equivalent to the filter estimation for the time-varying component $\tilde{\sigma}_c^2(t)$ as well.

If the elements of $\boldsymbol{R}_{xx}(t,\tau)$ are estimated using a time frame of length $N$ which is significantly larger than the period $T$. In this scenario, where $N \gg T$, the time-dependent function $\sigma_c^2(t)$ can be represented as the sum of the average $\bar{\sigma}_c^2$ and the time-varying component $\tilde{\sigma}_c^2(t)$ as follows:

$$\boldsymbol{R}_{xx}(t,\tau) \underset{N \gg T}{=} \underbrace{(\bar{\sigma}_c^2 + \sigma_w^2)\boldsymbol{\Phi}(\tau)}_{\bar{\boldsymbol{R}}_{xx}(\tau)} + \underbrace{\tilde{\sigma}_c^2(t)\boldsymbol{\Phi}(\tau)}_{\tilde{\boldsymbol{R}}_{xx}(t,\tau)}$$

$$\bar{\sigma}_c^2 = \frac{1}{N}\sum_{t=1}^{N}\sigma_c^2(t), \quad (10)$$

where $\bar{\sigma}_c^2$ is the average of $\sigma_c^2(t)$ over time frame $N \gg T$.

The deconvolution filter, constructed using $\bar{\boldsymbol{R}}_{xx}(\tau)$ as in the first part of equation (10), will be equivalent to the filter required by the second part of the equation. This is because both parts depend on the matrix $\boldsymbol{\Phi}(\tau)$, which is solely defined by the coefficients of the constant TF that the deconvolution filter aims to remove.

Therefore, a constant deconvolution filter exists and can be defined as $\boldsymbol{g}^T = [g_1, g_2, \ldots g_n]$. This filter is used to convert $x(t)$ into a zero-mean and uncorrelated signal $y(t)$, where the length of $y(t)$ is assumed to be equal to the length of the TF. The conversion is as follows:

$$y(t) = \sum_{i=1}^{n} g_i x(t-i) \cong c(t) + w(t). \quad (11)$$

Here, $n$ is assumed to be the length of $\boldsymbol{g}$. However, in general, it can differ from the length of the TF.

After estimating $\boldsymbol{R}_{xx}(t,\tau)$ using $N \gg T$ and obtaining $\bar{\boldsymbol{R}}_{xx}(\tau)$, which is constant over time, we employ the Mahalanobis whitening approach [25] to construct the transformation $\boldsymbol{g}$. This is done by converting $\bar{\boldsymbol{R}}_{xx}(\tau)$ into a diagonal matrix, while keeping the signal energy unchanged:

$$\boldsymbol{R}_{yy}(\tau) = \boldsymbol{G}\boldsymbol{R}_{xx}(\tau)\boldsymbol{G}^T = \bar{\sigma}_x^2 \boldsymbol{I}$$

$$\boldsymbol{G} = \bar{\sigma}_x \bar{\boldsymbol{R}}_{xx}^{-1/2}(\tau), \quad (12)$$

where $\boldsymbol{G}$ and $\boldsymbol{I}$ are the $n \times n$ whitening transformation and the unit diagonal matrices respectively.

Matrix $\boldsymbol{R}_{yy}(\tau)$ is a diagonal matrix with identical elements along its diagonal. Consequently, each column of the matrix $\boldsymbol{G}$ signifies a valid solution, with the only difference being the phase shift. The central column of $\boldsymbol{G}$ provides a zero-phase solution. Therefore, if n is an odd number, the desired filter will be equivalent to this solution:

$$\hat{\boldsymbol{g}} = \boldsymbol{G}\boldsymbol{e}_{(n-1)/2}, \quad (13)$$

where $\boldsymbol{e}_{(n-1)/2}$ is the $n \times 1$ basis vector with 1 at $(n-1)/2$ row and zero otherwise.

*4.2. CS2 Envelope Estimation*

The CS2 component, as defined in (2), is given by: $c(t) = q(t)r(t)$, where $q(t)$ represents a non-negative periodic envelope of interest, and $r(t)$ is a zero-mean white noise. As shown in (12), $y(t)$ is a zero-mean, uncorrelated variable, the autocorrelation function of which can be expressed as follows:

$$r_{yy}(t, \tau) = \mathbb{E}\{y(t)y(t - \tau)\} = \mathbb{E}\{y(t)^2\}\delta(\tau) = (q^2(t)\sigma_r^2 + \sigma_w^2)\delta(\tau), \tag{14}$$

where $\delta(\tau)$ is a delta of Dirac.

The variance of $r(t)$ is a constant value. We can set this value to $\sigma_r^2 = 1$ without loss of generality, given that the waveform of $q(t)$ is of our primary focus. If the variance $\sigma_w^2$ is known, then in theory, we can estimate $q(t)$ by subtracting $\sigma_w^2$ from the $\mathbb{E}\{y(t)^2\}$:

$$\hat{q}(t) = \sqrt{\mathbb{E}\{y(t)^2\} - \sigma_w^2}. \tag{15}$$

In practical scenarios, the estimation of $\mathbb{E}\{y(t)^2\}$ requires a time window that is significantly smaller than the period of $q(t)$ in order to detect its temporal variations. However, given that the period of $q(t)$ is unknown an alternative solution is needed.

We suggest estimating it in the frequency domain by leveraging the periodic characteristics of $q(t)$. Assuming the Fourier transform of $y(t)^2$ exists, the expression $\mathbb{E}\{y(t)^2\}$ can be reformulated as follows:

$$\mathbb{E}\{y(t)^2\} = F^{-1}(\mathbb{E}\{F([y^2(t)])\}) \cong F^{-1}\left(\mathbb{E}\{|F(y^2(t))|\}e^{-j\angle F(y^2(t))}\right), \tag{16}$$

where operators $F()$ and $F^{-1}()$ denote the Fourier transform and the inverse Fourier transform, respectively; the operators $|*|$ and $\angle$ correspond to the amplitude and phase values, respectively. The expectation $\mathbb{E}\{|F(y^2(t))|\}$ can be approximated by estimating the amplitude using the Welch periodogram and transitioning back to the time domain using a single frame phase. In this scenario, the estimation of $\mathbb{E}\{y(t)^2\}$ does not require information about the expected period of $q(t)$, as the expectation is estimated in the frequency domain.

If the variance $\sigma_w^2$ is unknown, the estimation of $q(t)$ in (15) will exhibit bias. If $q(t)$ comprises a limited number of cyclic components, the estimation variance of $\mathbb{E}\{y(t)^2\}$ can be minimized by identifying peaks in the spectrum and eliminating frequencies with peak amplitudes below a certain threshold before transitioning back to the time domain. The threshold can be estimated by the median of $|F(y^2(t))|$, under assumption that only limited number of peaks in the spectrum are associated with $q(t)$.

*4.3. Algorithm Summary*

**Input**:
Receive $N$ measurements $\{x(1), \ldots x(N)\}$ as defined in (5):

$$x(t) = \sum_{i=1}^{n} h_i y(t - i)$$

$$y(t) = q(t)r(t) + w(t)$$

**Parameters:**

$n$ is odd and corresponds to the length of the TF impulse response $\boldsymbol{h}$ and the inverse TF $\boldsymbol{g}$.
$L \leq N$ is the Fourier transform window and is utilized in the Welch periodogram.

**Required Output:**
Estimated envelope $\hat{q}(t)$.
**The Algorithm:**
1. Estimate $n \times n$ correlation matrix $\boldsymbol{R}_{xx}$ by using $N$ measurements:

$$R_{xx} = \begin{bmatrix} r_x(0) & \cdots & r_x(1-n) \\ \vdots & \ddots & \vdots \\ r_x(n-1) & \cdots & r_x(0) \end{bmatrix}$$

$$r_x(\tau) = \frac{1}{N}\sum_{t=1}^{N} x(t)x(t-\tau), \tau = 0,1,\ldots,n$$

2. Estimate whitening transformation $G$:

$$G = \left(\sum_{t=1}^{N} x(t)^2 \, R_{xx}^{-1}\right)^{1/2}$$

3. Estimate the deconvolution filter as a middle column of $G$:

$$\hat{g} = G e_{(n-1)/2}$$

4. Perform deconvolution of $x(t)$:

$$y(t) = \sum_{i=1}^{n} g_i x(t-i), \quad t = 1,\ldots N$$

5. Estimate $\mathbb{E}\{|F[y^2(t)]|\}$ by employing the Welch periodogram of $y^2(t)$ with a window size of $L$. If $q(t)$ is presumed to contain a finite number of cyclical components, retain only the peaks that exceed the median of $|F[y^2(t)]|$. Utilize the phase information from a single frame and implement the inverse Fourier transform:

$$\mathbb{E}\{y(t)^2\} \cong F^{-1}\left(\mathbb{E}\{|F(y^2(t))|\} e^{-j\angle F(y^2(t))}\right)$$

6. If $\sigma_w^2$ is known, then utilize it. Otherwise set $\hat{\sigma}_w^2 = \min(\mathbb{E}\{y(t)^2\})$ to ensure that the minimum of $\hat{q}(t)$ is zero. In this scenario the estimation will be biased:

$$\hat{q}(t) = \sqrt{\mathbb{E}\{y(t)^2\} - \sigma_w^2}$$

## 5. Experimental Results

*5.1. Simulations*

The simulations were designed to assess the robustness of the new method with respect to three parameters: the number of poles in the TF, the number of cyclic components in $q(t)$ and the noise $w(t)$.

Each combination of parameters was subjected to 100 iterations of the algorithm. The variance of noise $w(t)$ was treated as a known quantity, resulting in unbiased estimation.

Signals $x(t)$ each lasting one second and consisting of $N = 24,000$ samples were generated during each run using (5) as follows:

$$x(t) = \sum_{i=1}^{n} h_i\big(q(t-i)r(t-i) + w(t-i)\big), t = 1,\ldots,N$$

$$q(t) = \sum_{k=1}^{K} \big(1 + B_k \cos(\omega_k t + \varphi_k)\big), \quad r(t) \sim \mathcal{N}(0,1), \quad w(t) \sim \mathcal{N}(0, \sigma_w^2).$$

The simulation settings were adjusted prior to each run as follows:

- The number of poles in the TF, which define the coefficients $h_i$ and their length $n$, varied between 5 and 20. These poles were uniformly distributed around the unit circle.
- The number of cyclic components $K$ in $q(t)$ ranged between 5 to 20. The frequencies $\omega_k$ and phases $\varphi_k$ were uniformly distributed within the range $(-\pi, \pi)$.
- The CS2 noise variance of $r(t)$ was set to 1.
- The variance of the zero-mean Gaussian noise $w(t)$ was determined based on the required SNR and signal energy.
- The SNR, defined as $SNR = 10 \log 10 \frac{\|q(t)\|_2^2}{\sigma_w^2}$, was specified separately for each run and varied between -20dB and 20dB.

The performance of the new method was evaluated using the Coefficient of Determination $R^2$ as follows:

$$R^2 = 1 - \frac{\sum_{t=1}^{N}(q(t) - \hat{q}(t))^2}{\sum_{t=1}^{N}(q(t) - \bar{q})^2}, \qquad (17)$$

where $N = 24{,}000$ is the length of the signal, $q(t)$ is the true value, $\hat{q}(t)$ is the estimated value, $\bar{q}$ is the average of the true signal.

An example of the generated signal $x(t)$ is demonstrated in Figure 1. The component $q(t)$ consists of $K = 6$ cycling components with SNR of -5dB and the TF has 8 poles.

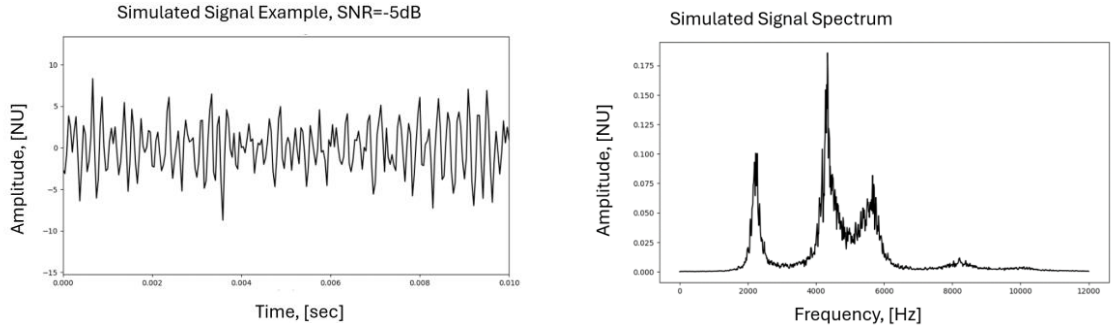

Figure 1. An example of the simulated signal $x(t)$ (on the left) and its corresponding spectrum (on the right) is provided. The signal $q(t)$ comprises $K = 6$ components, and the TF incorporates 8 poles.

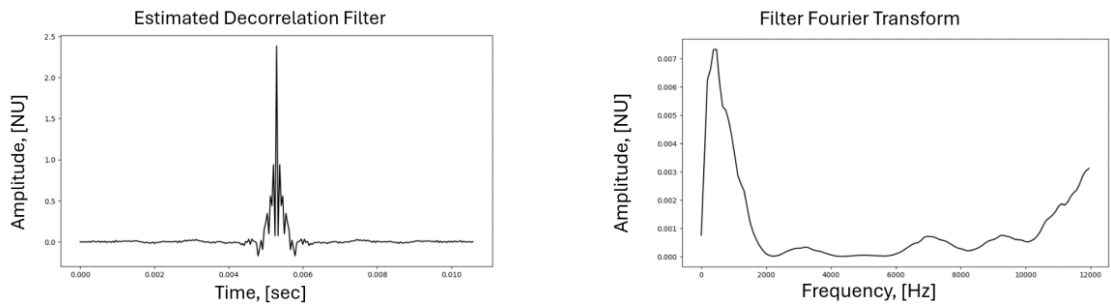

Figure 2. An example of the deconvolution filter waveform (on the left) and its corresponding Fourier transform (on the right).

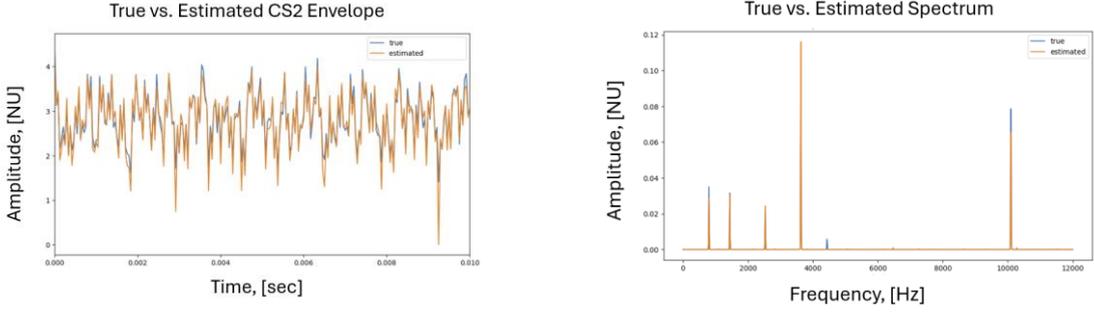

Figure 3. An example of estimated component, $q(t)$ (on the left) and its spectrum (on the right) under SNR of -5dB. The true signal is represented in blue, while the estimated signal is depicted in orange.

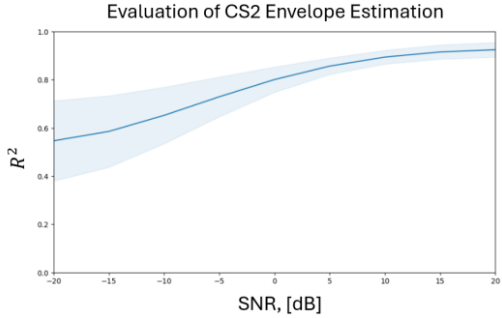

Figure 4. The relationship between the Coefficient of Determination $R^2$ vs. SNR, which is used to evaluate the estimation of the CS2 envelope, $q(t)$. The dispersion around the graph signifies a range of ±1 standard deviation.

*5.2. Error Analysis, Parameters Choice, and Method Limitations*

The simulations validate the method's robustness across various SNR conditions. Under low SNRs, the estimation variance is notably higher and diminishes as the SNR increases. The outcomes correspond to diverse combinations of TF poles and cyclic components in $q(t)$, indicating that the method's performance is independent of signal parameters. There are two critical parameters to be determined: the length $n$ of the TF $\boldsymbol{h}$ and the deconvolution filter $\boldsymbol{g}$, and the frame length $L$ for Welch periodogram estimation. The length $n$ embodies the trade-off between the estimation accuracy, defined by the deconvolution filter estimation, and the computational cost. A high $n$ encompasses the required filter length and its frequency resolution but concurrently escalates computational complexity by enlarging the size of $\boldsymbol{R}_{xx}$. The inversion of this enlarged $\boldsymbol{R}_{xx}$ becomes more computationally intensive.

The selection of the frame length $L$ represents a balance between estimation variance and bias. The estimation variance will decrease by a factor of $L/N$. For instance, if $L/N = 10$, the estimation variance will decrease tenfold. Conversely, the frequency resolution deteriorates when $L$ decreases, and some cyclic components of $q(t)$ are not estimated, which escalates the estimation bias. The method is confined to a specific type of CS2 signals and a particular TF, which must be invertible. The CS2 signals are restricted to the multiplication between a periodic function and white noise. In practical scenarios, the measurement setup may also incorporate additional white noise added after the signals traverse the TF. In such cases, the noise may be amplified when the deconvolution filter is applied, necessitating further work to rectify this issue.

## 6. Conclusions

In this study, we presented an innovative method for dual estimation: the time waveform of second-order cyclostationary (CS2) signals and the deconvolution filter that mitigates the impact of the TF. This method, unlike its predecessors, does not require prior knowledge about the signals or TF, thereby broadening its practical applicability. It demonstrates high robustness to noise and remains unaffected by variations in signal and TF parameters. However, its application is restricted to a specific type of CS2 signals, specifically those resulting from the multiplication of a periodic signal and white noise. Future research is needed to estimate CS2 from other practical measurements, such as CS2 in the presence of noise and dominant CS1 components, as found in numerical practical applications. Furthermore, it would be intriguing to assess the performance of this method in Machine Learning applications, particularly when training data is aggregated from identical units but with different TFs.

**Declaration of competing interest**

The authors declare that they have no known competing financial interests or personal relationships that could have appeared to influence the work reported in this paper.